\documentclass{article}
\usepackage[preprint,nonatbib]{neurips_2020}

\usepackage[utf8]{inputenc} 
\usepackage[T1]{fontenc}    
\usepackage{hyperref}       
\usepackage{url}            
\usepackage{booktabs}       
\usepackage{amsfonts}       
\usepackage{nicefrac}       
\usepackage{microtype}      

\usepackage[pdftex]{graphicx}
\graphicspath{{figs/}}
\usepackage{amsmath}
\usepackage{amsthm}
\usepackage{siunitx}
\theoremstyle{theorem}

\theoremstyle{definition}
\newtheorem{definition}{Definition}
\newtheorem{example}{Example}

\newcommand{\prob}{\mathbb{P}}
\newcommand{\Prob}[1]{\prob\left[{#1}\right]}
\newcommand{\EE}{{\mathbb{E}}}
\newcommand{\EX}[1]{\EE\left[{#1}\right]}
\DeclareMathOperator{\TopW}{{TOP}}
\DeclareMathOperator{\ECE}{{ECE}}
\DeclareMathOperator{\NLL}{{NLL}}
\DeclareMathOperator{\BS}{{BS}}
\DeclareMathOperator{\variance}{Var}
\DeclareMathOperator{\Train}{{TRAIN}}
\DeclareMathOperator{\Test}{{TEST}}
\DeclareMathOperator{\KL}{{KL}}
\newcommand{\VX}[1]{\variance\left[{#1}\right]}
\DeclareMathOperator{\OR}{OR}
\DeclareMathOperator{\softmax}{softmax}

\title{Calibrated Top-1 Uncertainty estimates for classification by score based models}
\author{%
  Adam M Oberman \\
  McGill University \\
  \texttt{adam.oberman@mcgill.ca} \\
    \And
  Chris Finlay\\
  McGill University \\
  \texttt{christopher.finlay@mcgill.ca} \\
  \And
  Alexander Iannantuono\\
    McGill University \\
    \texttt{alexander.iannantuono@mail.mcgill.ca}\\
  \And
  Tiago Salvador\\
  University of Michigan\\
  \texttt{saldanha@umich.edu}
}

\begin{document}
\maketitle
\begin{abstract}
While the \emph{accuracy} of modern deep learning models has significantly improved in recent years, the ability of these models to generate \emph{uncertainty estimates} has not progressed to the same degree. Uncertainty methods are designed to provide an estimate of class probabilities when predicting class assignment.
  While there are a number of proposed methods for estimating uncertainty, they all suffer from a lack of \emph{calibration}: predicted probabilities can be off from empirical ones by a few percent or more.  By restricting the scope of our predictions to only the probability of Top-1 error, we can decrease the calibration error of existing methods to less than one percent.  As a result, the scores of the methods also improve significantly over benchmarks. 
\end{abstract}

\section{Introduction}
Machine learning classification models have made groundbreaking gains in accuracy in recent years, using deep neural networks.  However, our ability to predict when models make errors has lagged behind.  As models are increasingly deployed to make decisions with real-world impacts, it becomes increasingly important to assess model error.  

When machine learning models make a probabilistic prediction, the accuracy of the prediction must be assessed.  For example if a model predicts that an image is dog with confidence .8, then 80\% of this time this occurs, the image should be a dog.
Any gap between predicted and empirical probabilities is referred to as \emph{calibration} error.  Methods which predict probabilities for class membership, can have significant calibration errors, even in the simplest setting of in-distribution error prediction.

\subsection{Contributions}
 By  focusing only on predicting the probability that the classification is \emph{correct} on a given image, (Top-1), instead of predicting probabilities for all class labels, we are able to significantly reduce the calibration error of models, and improve the Brier scores compared to benchmark methods \cite{SnoekOFLNSDRN19}.  
  
\begin{figure*}
\begin{center}
  \centering
   \includegraphics[width=.6\paperwidth]{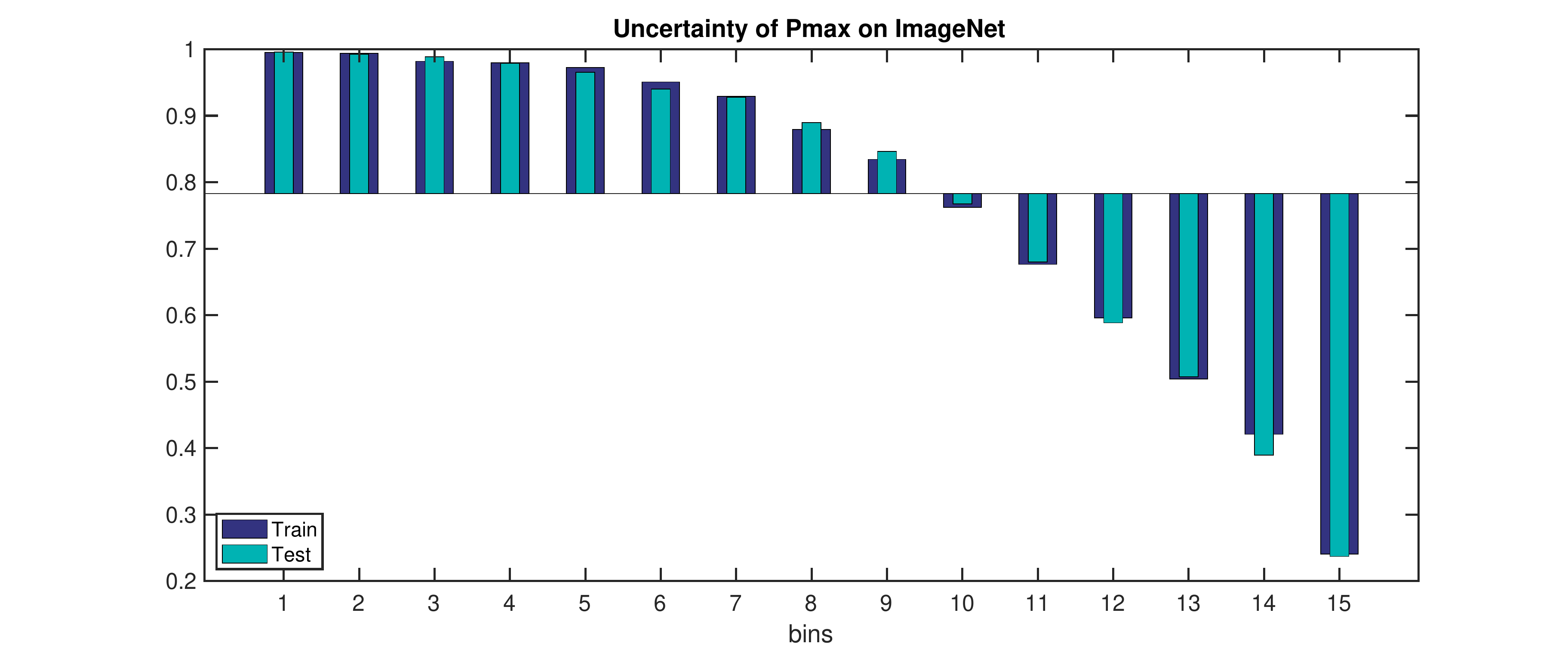}   
\caption{Predictions that image classification is correct (Top-1)  using the pre-trained ResNet-152 model the ImageNet dataset.  The values of Pmax are binned, based on the value of $Pmax$ uncertainty method \cite{hendrycks_baseline_2017}. Bins are designed to be equal size, and heights are plotted against the baseline Top-1 test accuracy of the model of 78.3\%.  The prediction is that images in the first bin will be correctly classified probability $.995 \pm .0025$, whereas images in the last bin will be correct with probability $.211 \pm .005$.
Test and training bins are superimposed, with height gap indicating the difference 
between test and training predictions. The Expected Calibration Error (ECE), which measures the average gap, is~0.69\%, which is significantly less than benchmark of 2-3\% \cite{SnoekOFLNSDRN19}.    }
\label{fig:calibration}
\end{center}
\end{figure*}

 Our method is designed for \emph{Top-1 accuracy} prediction, rather than also making   predictions for lower ranked classes.  By restricting the scope of predictions, we can obtain better calibrated predictions using existing uncertainty measures (e.g. softmax probability values (Pmax)~\cite{hendrycks_baseline_2017}).  The method is to simply bin the uncertainty values.  Bin accuracy is high, as illustrated in \autoref{fig:calibration}. \autoref{tab:brier} shows the  the improvements to scores using Top-1 binning: the expected calibration error (ECE) is reduced by a factor of three, and the Brier score (smaller is better), which measures the mean squared loss between the predicted probability and the Top-1 error, is improved by a factor of three on ImageNet, and three to five on CIFAR-10.  The implementations in \cite{SnoekOFLNSDRN19} for image classification have an expected calibration error (ECE) of several percent, for example, the ECE of the Vanilla method on CIFAR-10 was 5\%.

Binning~\cite{zadrozny2001learning} is a simple method for predictive  uncertainty. Binning methods yield conditional probabilities which are calibrated, up to the statistical error of sampling the bins.  While the statistical error can be significant in the multiclass setting, Top-1 predictions give a statistical error comparable to the binary classification case.   \cite{zadrozny2002} separates multiclass  problem into a number of binary problems, and calibrated the scores from each binary classifier, then combined them to obtain multiclass probabilities.   \cite{naeini2015} extends histogram binning to consider multiple binning models and their combination.  Our approach is a special case of the method in~\cite{KuleshovL15}, which sought to improve prediction errors by focusing on predicting only events of interest, which include the Top~1 error.  Histogram binning as implemented in \cite{guo_calibration_2017} led to a loss of accuracy in the multiclass setting.  
  
Binning allows us to obtain better calibrated Top-1 prediction estimates, using existing uncertainty measures.  The cost is that we no longer have predictions for lower ranked classifications.  However, the gains in accuracy of the predictions can make  the trade off worthwhile, in many instances.  Furthermore, we can also predict other events, such as Top-5 accuracy, using a very similar method. 

\begin{table*}
  \caption{Comparing the proposed Top-1 implementation of uncertainty measures with the benchmark implementation of the same measures. 
  Top: Brier score (smaller is better) versus benchmarks in~\cite{SnoekOFLNSDRN19}.  The Top-1 binning method scores better for each uncertainty measure.  
  Bottom: Expected Calibration Error (ECE) for Top-1, compared to the range of ECE for methods in \cite{SnoekOFLNSDRN19}.}
   \label{tab:brier}
\begin{center}
\begin{small}
\begin{sc}
  \begin{tabular}{ l     c c c c }    
    \toprule
Brier score    &  \multicolumn{2}{c}{CIFAR-10}  & \multicolumn{2}{c}{ImageNet-1K}   \\
\midrule
      uncertainty measure &  Top-1 & \cite{SnoekOFLNSDRN19}  & Top-1& \cite{SnoekOFLNSDRN19} \\
      \midrule
      Vanilla      	&  \textbf{0.033}  & 0.18 & \textbf{0.11~~}  & 0.34 \\
    Dropout 		& \textbf{0.036}  &0.17  & \textbf{0.135} & 0.35  \\
    Ensemble variance     & \textbf{0.040} &  0.11 & \textbf{0.135} & 0.31  \\
    \bottomrule
  \end{tabular}
 \vskip 0.1in 
    \begin{tabular}{ l c c}
        \toprule
        Expected Calibration Error      & CIFAR-10   & ImageNet   \\ \hline
         Top-1                        & \textbf{1.-1.1\%} & \textbf{0.9-1.2\%} \\ 
        \cite{SnoekOFLNSDRN19}  &  1.0-5.0\%     & 2.0-3.0\%      \\ 
        \hline
    \end{tabular}
  \label{tab:cifar100-bins}
 \end{sc}
\end{small}
\end{center}
\end{table*}

%
%
%

\section{Background}

\subsection{Classification and label probabilities}
Let $y \in \mathcal Y = \{ 1, \dots, K\}$ be a label from one of $K$ classes.
Consider a model $f(x) = (f_1(x), \dots f_K(x))$ which outputs a score function $f_i$ for each class $i$.
The classification of the model corresponds to the highest score $
\hat y(x) =  \arg\max f_i(x)$.   
Let $y(x)$ be the correct label of $x$, we say the model is Top-1 correct if $\hat y(x)=y(x)$.
Predictive uncertainty seeks to estimate the probability that an image belongs to each class,
\begin{equation}\label{confidence_label_i}
	p^{class}_k(x) = \Prob{ y = k \mid x}, \qquad \text{ for each } k \in \mathcal Y
\end{equation}
The softmax function transforms model outputs  into a probability vector,
$p^{f}_i(x) = \softmax(f(x))_i$.
The softmax values $p^f$ are sometimes confused with $p^{class}_i(x)$, but generally speaking,  $p^f_i(x)$ is not an accurate prediction of $p^{class}_i(x)$~\cite{domingos1996beyond}. 

Consider instead to the rank-$k$ event, that the $k$-th largest score function corresponds to the correct label.  Let $\hat y^{rank}_{k}$ be the index of $k$-th largest score function, and consider the probability of Rank-$k$, 
\begin{equation}
	\label{confidence_top_i}
	p^{rank}_k(x) = \Prob{ y = \hat y^{rank}_k \mid x}, \qquad \text{ for each } k \in \mathcal Y
\end{equation}

In this work, we focus on estimating  the Top-1 probability (which is the same as $p^{rank}_1$), which we write as 
$p^{\TopW} = p_1^{\TopW}= p^{rank}_1$.  In the case of ImageNet, we also estimate and $p^{TOP}_5 = \sum_{i=1}^5 p^{rank}_k$.  We estimate the probability using binning, described below. 

\subsection{Predictive uncertainty methods} 
In the setting of deep neural networks, there are a number of proposed approaches 
for estimating \eqref{confidence_label_i}. We refer to the recent work \cite{SnoekOFLNSDRN19} which compares a number of uncertainty methods, including 
\begin{itemize}
	\item (Vanilla) Maximum softmax probability (PMax)~\cite{hendrycks_baseline_2017}
\item (Dropout) Monte-Carlo Dropout \cite{gal2016dropout} \cite{srivastava2014dropout} with rate $p$. Dropout \cite{srivastava2014dropout}  was interpreted in a Bayesian setting by \cite{gal_uncertainty_2016} \cite{gal2016dropout} and \cite{kingma_variational_2015} This interpretation was later retracted \cite{hron_variational_2018}), however it remains an effective empirical method. 
\item  (Ensembles) Ensembles of $M$ networks trained independently on the entire dataset using random initialization. Model ensembles \cite{lakshminarayanan_simple_2017} use multiple models to make a classification: the uncertainty estimate is based on the degree of agreement between the predictions \cite{Dietterich00,li2018survey}.  
\item (Temp Scaling) Post-hoc calibration by temperature scaling using a validation set \cite{guo_calibration_2017}
\end{itemize}

Our implementation of these methods involves choosing a scalar (single numeric) quantity $U(x)$ coming from the choice of method, and then binning the values.  PMax is already a scalar value.  Dropout and the ensemble method return $M$ vectors $f_1, \dots, f_M$.   In our implementation,  we combine the output into a single number: the matrix 2-norm of the covariance matrix $\| \variance(f_i)\|_2$.   The values are then binned as described below. Temperature scaling is a transformation of the output of the model which does not change the bins for our method, compared to the original model.  So we do not report results for temperature scaling.

\section{Calibrated Top-1 Uncertainty bins} \label{sec:bins}

We restrict the scope of our predictions to just the event that the classification of the model is \emph{correct}, in other words, the Top-1 error, \eqref{confidence_top_i}, or the Top-5 error.  

\begin{figure*}
\vskip 0.2in
\begin{center}
  \centering
  {\includegraphics[width=.3\paperwidth]{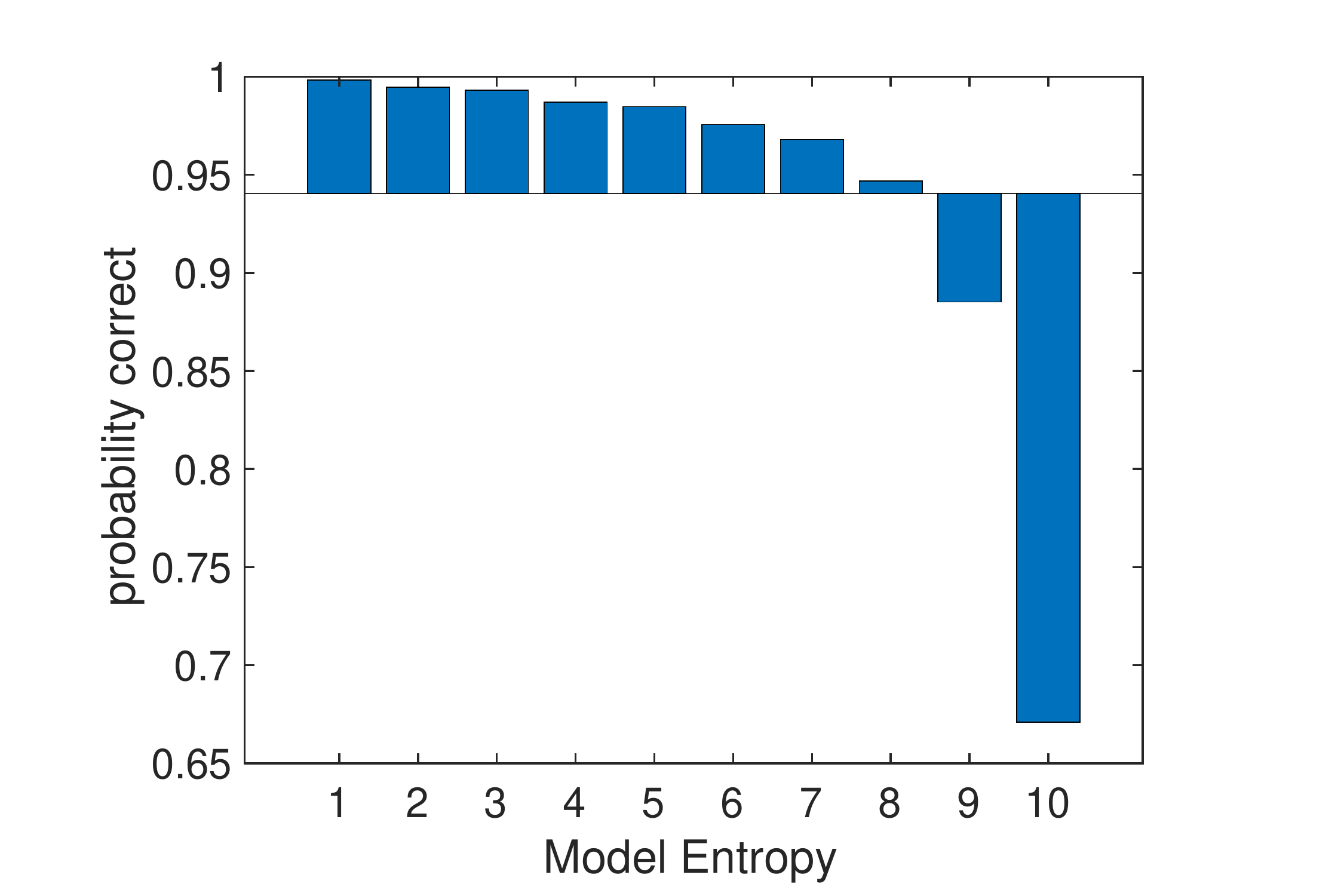}\includegraphics[width=.3\paperwidth]{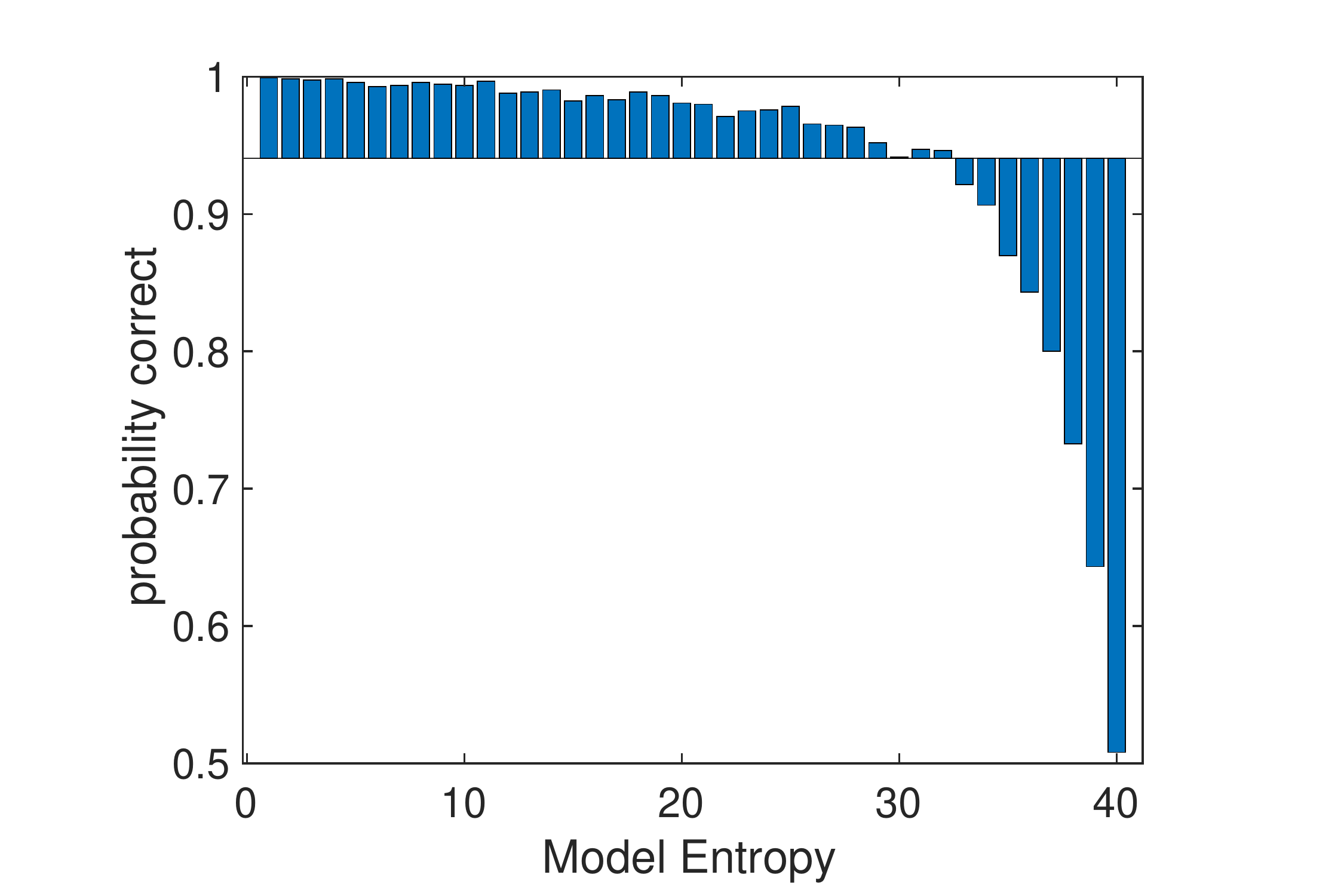}}
\caption{Using more bins better approximates the conditional probability.} 
\label{figRefine2}
\end{center}
\vskip -0.2in
\end{figure*}

Write $\TopW(x) = 1_{\hat y = y}$ the Top-1 random variable  ($\TopW(x) =1$ if the classification of the model  is correct, and $0$ otherwise).  Let $U$ be an uncertainty random variable.   We define the conditional probability coming from a binning process as a random variable. 
\begin{definition}[Uncertainty bins]\label{defn:Conf}
Given  the uncertainty random variable $U(x)$, and the bins $B_i$ with edges $e_i, e_{i+1}$, define the prediction random variable, $C$ as the conditional probability 
 \begin{equation}\label{U_Cond_Prob}
 \begin{aligned}
 	p^{\TopW}_i &= \EX{ \TopW(x) \mid  U(x) \in (e_i, e_{i+1}]~ },
 \end{aligned}
 \end{equation}
 with weights  $w_i = \Prob{U \in B_i }$.  Write $\bar p = \EX{\TopW }$ for the accuracy of the model. 
\end{definition}
The random variable, $C$, is a conditional probability, so it is calibrated, by definition~\eqref{U_Cond_Prob}. Given the bins, $B_i$, determined by the edges, and a test or training data set, We write $
C^{\Train} = (w_i^{\Train}, p_i^{\Train}),
C^{\Test} = (w_i^{\Test}, p_i^{\Test})$
for the test and training estimates of the prediction variable $C$, which are defined as follows. 
 \begin{itemize}
 	\item Evaluate $U$ and $\TopW$ and bin the predictions of the values of $U(x)$ using the edges $E$, according to \eqref{U_Cond_Prob}. 
  \item  Let $|B_i|$ be the number of points in each bin. The empirical weights and predicted confidence  of the bin are given by  
\[
\hat p_i =  \frac{1}{|B_i|} \sum_{j \in B_i} \TopW(x_j), \qquad
w_i = \frac{|B_i|}{\sum_{k=1}^M |B_k|}
\]
 \end{itemize}
In practice, we use equal sized bins, determined by the quantiles of the training data, in order to minimize the empirical calibration error $|p_i^{\Train}- p_i|$ of each bin, which is of order $1/\sqrt{|B_i|}$.  More precisely, for any $\delta > 0$, 
\begin{equation}
	\label{interval_bound}
	| p_i - p^{\Test}_i| \leq \sqrt{\frac {\log ( 2 / \delta)}{2 |B_i|} }, \qquad \text{ with probability } \geq 1- \delta.
\end{equation}
 The result follows from Hoeffding's inequality in the case of Bernoulli random variables \cite[Chapter 1]{wasserman2006all}.    In \autoref{figRefine2} we show how increasing the bins approximates the true conditional probability $C(x)$.   More sophisticated estimates of the conditional probability could be performed using kernel density estimation, with bounds for the probability density, as in \cite{wasserman2006all}.

\section{Scoring the predictions}
\subsection{Proper scoring rules: Brier and NLL}
 Proper scoring rules~\cite{gneiting_strictly_2007} \cite[Chapter 10]{parmigiani2009decision} are used to score predictive uncertainty estimates. 
 In our setting, calibration error measures how far empirical bin probabilities $\hat p_i$ are from the true values $p_i$, given by~\eqref{U_Cond_Prob}.    Resolution measures how far the values $p_i$ are from the accuracy $\bar p$. 
  Scoring rules can be decomposed into terms involving calibration error and  resolution, see \S\ref{sec:decompose}.

In the binary case, the Brier score~\cite{brier_verification_1950} is given by 
\[
\BS(\TopW,C) 
= \EX { (p(x)-\TopW(x))^2  }
\]
In this case, the resolution is the variance of $C$, and the calibration error is the $\sum w_i |p_i - \hat p_i|^2$. 

The  negative Log-Likelihood (NLL)  score \cite{goodj} is given by 
\[
{\NLL}(\TopW, C) = \EX {-\log(p(x))\TopW(x) -\log(1-p(x))(1-\TopW(x)) }
\]

\subsection{Expected Calibration error}
The Expected Calibration error (ECE) is a popular method for measuring calibration error.  The  ECE between an empirical prediction variable $\widehat C$ and the true prediction random variable $C$ defined by \eqref{U_Cond_Prob} is given by 
\[
\ECE(C,\widehat C) = \sum_i  w_i |p_i - \hat p_i|.
\]
The ECE is approximated using empirical test and training histograms by 
\begin{equation}\label{CE}
\ECE(C^{\Test},C^{\Train}) = \sum_i   w^{\Test}_i |p^{\Test}_i - p^{\Train}_i|.
\end{equation}
 A large ECE is always bad, but a zero ECE is not necessarily good. For example, always predicting $\bar p$ gives an ECE of zero but has zero resolution. 

\subsection{Odds ratio as a measure of resolution}
Given a probability $p \in (0,1)$, the odds of $p$ is $O(p) = p/(1-p)$. 
We propose the odds ratio as a measure of resolution. 
\begin{definition}[Odds Ratio]\label{defn:EOR}
For $p,a \in (0,1)$ define
\[
	\OR( p, a ) =  \max \left (  \frac{O(p)}{O(a)},    \frac{O(a)}{O(p)} \right ) 
\]
For a prediction random variable, $C$, the Expected Odds Ratio is 
\[
\mathrm{EOR}(C) = \sum_i w_i \OR(p_i, \bar p)
\]

\end{definition}
The EOR is a relative measure of the resolution of $C$.  It is interpreted as 
 the amount we expect our winnings to increase, if we could make bets with odds $O(\bar p)$, given the information provided the probability bins in~\S\ref{sec:ORexplained}.

\section{Scoring and decompositions}\label{sec:decompose}

\subsection{Brier and NLL scores decomposed}
\cite{degroot1983comparison} studies probabilistic scoring methods in the binary case, in particular the decomposition of scoring rules into calibration error and resolution. 
This decomposition was generalized to multiclass~\cite{brocker2009reliability}.

Let $C$ be the exact  prediction random variable and let $\widehat C$ be an empirical prediction random variable. For the Brier score, we have the decomposition
\[
\BS(\TopW,\widehat C) = \underbrace{\bar p(1-\bar p)}_{\text{uncertainty}}
- \underbrace{\sum w_i(\bar p - p_i)^2}_{\text{resolution}}
 + \underbrace{\sum w_i ( p_i - \hat p_i)^2}_{\text{calibration error}}
\]
The resolution is just the variance of the accuracy over the bins, 
which is the width of the centered bins in \autoref{fig:calibration}. The calibration error is the mean-squared difference of confidence and accuracy over the bins.  Thus the Brier score is minimized if (i) calibration error is zero and (ii) the resolution is as large as possible (equal to the uncertainty).   

The decomposition of the NLL score is given by
\[
\NLL(\TopW,\widehat C(x)) = 
\underbrace{ \mathrm H(\bar p)}_{\text{uncertainty}}
- \underbrace{\sum w_i \KL(\bar p, p_i)}_{\text{resolution}}
 + \underbrace{\sum w_i \KL(\hat p_i,p_i)}_{\text{calibration error}}
\]
where the entropy is $\mathrm {H}(p) =\sum -p_i \log(p_i)$
and the divergence term is given by the KL divergence $\KL(p,q) = - \sum q_i \log \left( p_i / q_i \right)$.  Again, we have an interpretation of the terms: entropy is an information theoretical measure of the uncertainty, and the KL divergence terms are measures of spread. 
 
The derivation of the decomposition of the Brier score is as follows (see, for example, \cite{KuleshovL15})
\begin{align*}
	\BS(\TopW,\widehat  C) & = \EX{(\TopW - \widehat  C)^2 } = \EX{(\TopW -  C)^2 } +  \EX{( C- \widehat C)^2 }\\
    &= \underbrace{\VX{\TopW}}_{\text{uncertainty}}
	 -\underbrace{\VX{C(x)}}_{\text{resolution}} +  
	 \underbrace{\EX{(\widehat C (x)- C(x))^2 }}_{\text{calibration error}} 
\end{align*}

\subsection{Explaining the Odds Ratio resolution}\label{sec:ORexplained} \begin{example}\label{ex:coin}
Consider the experiment consisting of tossing one of four randomly chosen coins.  Coins 1 and 2 are fair, while coins 3 and 4 have  $p(H) = 15/16$ and  $1/16$, respectively.   Clearly, $p(H) = .5$ for this experiment. 
Next, suppose we can identify the coins, and let $U \in \{1,2,3,4\}$ represent the chosen coin.  Conditioning on the value of $U$, we have the following histogram:  $p = (1/2, 1/2, 15/16, 1/16)$, $w = (.25, .25, .25, .25)$.  Knowing which coin is being tossed changes the odds of heads from 1:1 to 15:1 or to 1:15.
\end{example}

The odds for the baseline probability are $O(a) = a/(1-a)$.  If new information changes the probability to $p$, the new odds are $O(p) = p/(1-p)$. The value of the new information must be compared to the odds.   The odds ratio measures how much the expected winnings of a fair bet increase, when new information is available.
When $p \geq a$ are given by the odds ratio is $\OR( p, a ) =     \frac{O(p)}{O(a)}$.  On the other hand,   if the odds have decreased, we should bet against, so the odds ratio is $\OR( p, a ) =     \frac{O(a)}{O(p)}$. 
 
The EOR for the coin toss Example~\ref{ex:coin} is 8.  This value corresponds to the fact that for a single bet, we expect to win 15 each time the biased coins are tossed, and 1 each time the fair coins, so the EOR is (15 + 15 + 1 + 1)/4 = 8.

\section{Empirical results}

In our empirical results, we focus on how improving calibration error results in better scores, using the Brier score and the NLL score.  Calibration error is measured by the ECE.  Since the calibration error is nearly the same for all methods, the methods can also be compared using the EOR measure of resolution.

\subsection{Methodology}

In practice, we start with a calibration (test) set, choose the number of bins, $M$,  and define the bin edges using quantiles, so that there are an equal number of points in each bin.
We chose bin sizes so that our ECE was always less that one percent, which in practice was consistent with the $1/\sqrt{|B_i|}$ estimate above. 
We chose to calibrate using half the test set, and measure the error using the other half. 10-fold cross validation lead to ECE of < 1\% with a standard deviation < 0.1\% using on the order of 10-20 bins.  At 100 bins, the ECE was slightly over 1\%.   The choice of bin size did not affect the scores too much: the change to the scores was less that the differences of scores between different methods.  

In order to avoid probabilities of $0$, or $1$, we regularized the probabilities as follows: we set $\hat p^\epsilon_i =  \frac{|B_i|\hat p_i +  p_A}{|B_i|+1}$, where $p_A$ is the accuracy of the model.  This regularization, which corresponds to adding a single element to the bin with value $p_A$, can be interpreted as using a Bayesian prior with mean $p_A$ for the bin.  The regularization made a difference only in the case of  the CIFAR-10H human annotated data \cite{peterson2019human}, which otherwise had empty bins, due to the clustering of probabilities from human rankings. 
 	  
We have enough data so that the bin sizes allow us to use the bound \eqref{interval_bound} to obtain useful estimates on the error in our predictions.  For example, in the results which follow in \S\ref{sec:CIFAR10_results},  combining the first five (of ten total) bins for Dropout .02, we obtain $p^{\Test}_1 =  0.9983$ with $|B_1| = 2500$ which leads to a useful confidence interval:
with $\delta = .005$ we obtain $p_1 \geq 0.9643$ with probability at least .995. 

\subsection{Uncertainty on CIFAR-10}\label{sec:CIFAR10_results}
We compare different uncertainty methods on CIFAR-10: Pmax, Dropout, and data from CIFAR-10H \cite{peterson2019human}.
On the human classified images, we used the entropy of the human probability vector for each image.  All methods except dropout with variance .005 were effective.  Human scoring was better than the weak dropout, but not as effective as the other methods. 

\begin{figure*}
\vskip 0.2in
\begin{center}
  \centering
   {\includegraphics[width=.33\paperwidth]{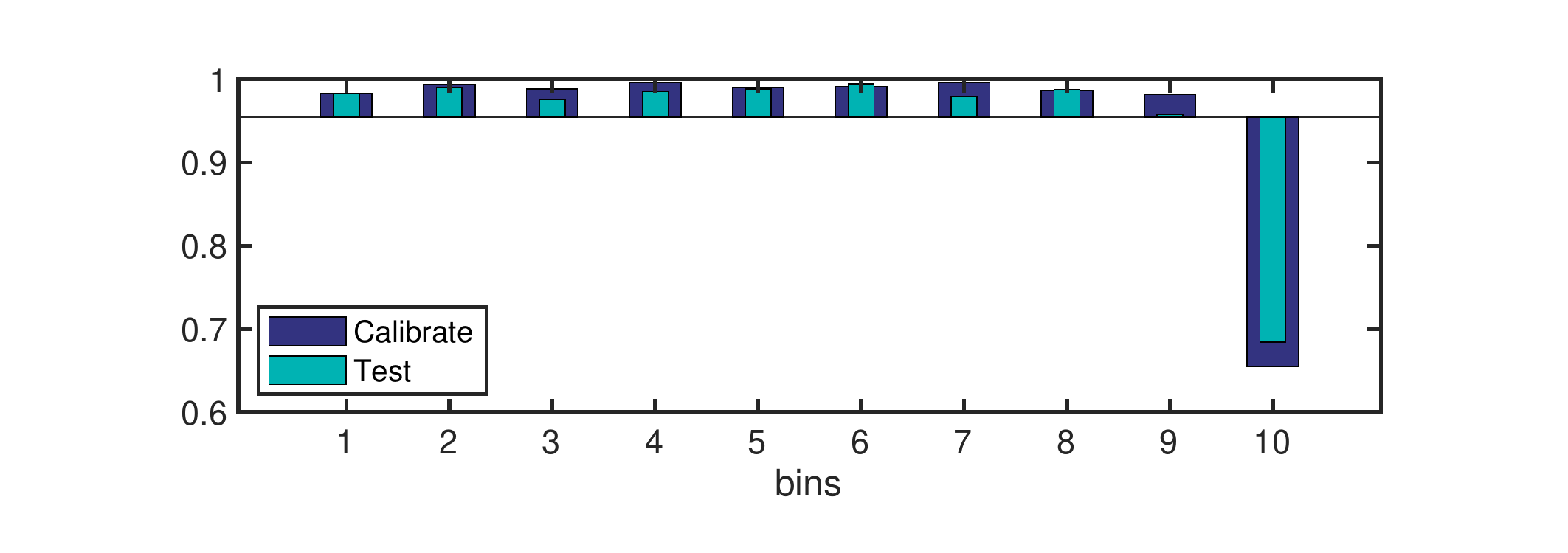}\includegraphics[width=.33\paperwidth]{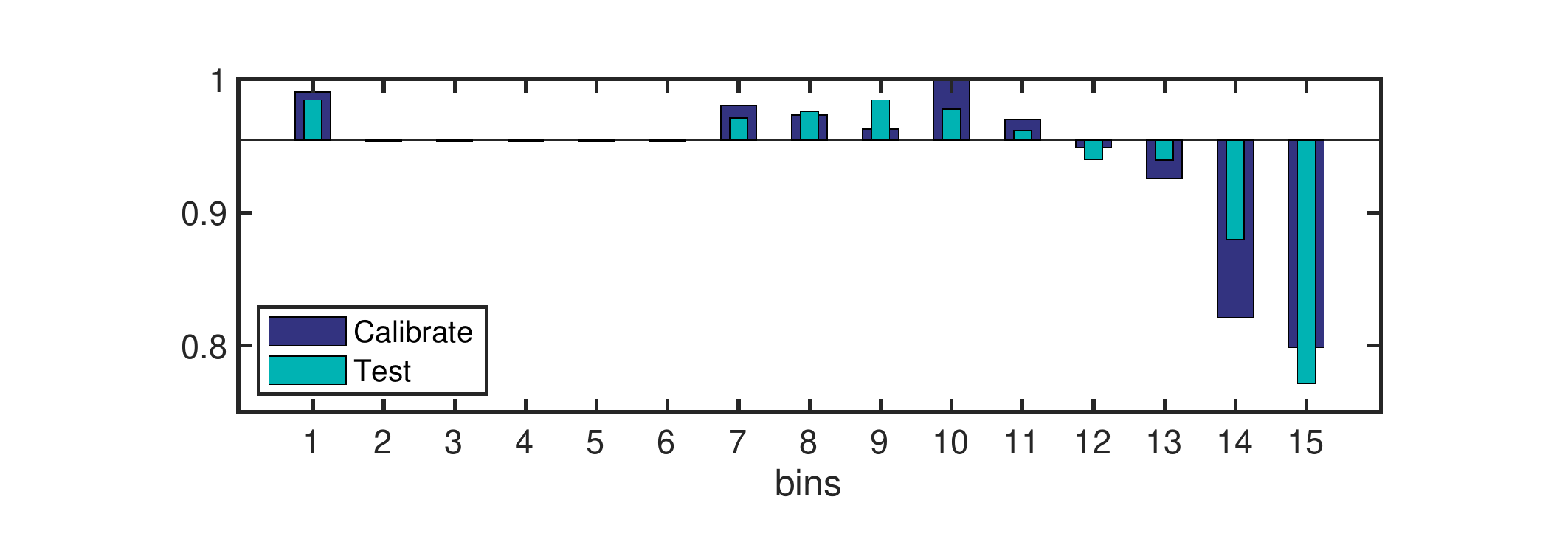}}
  { \includegraphics[width=.33\paperwidth]{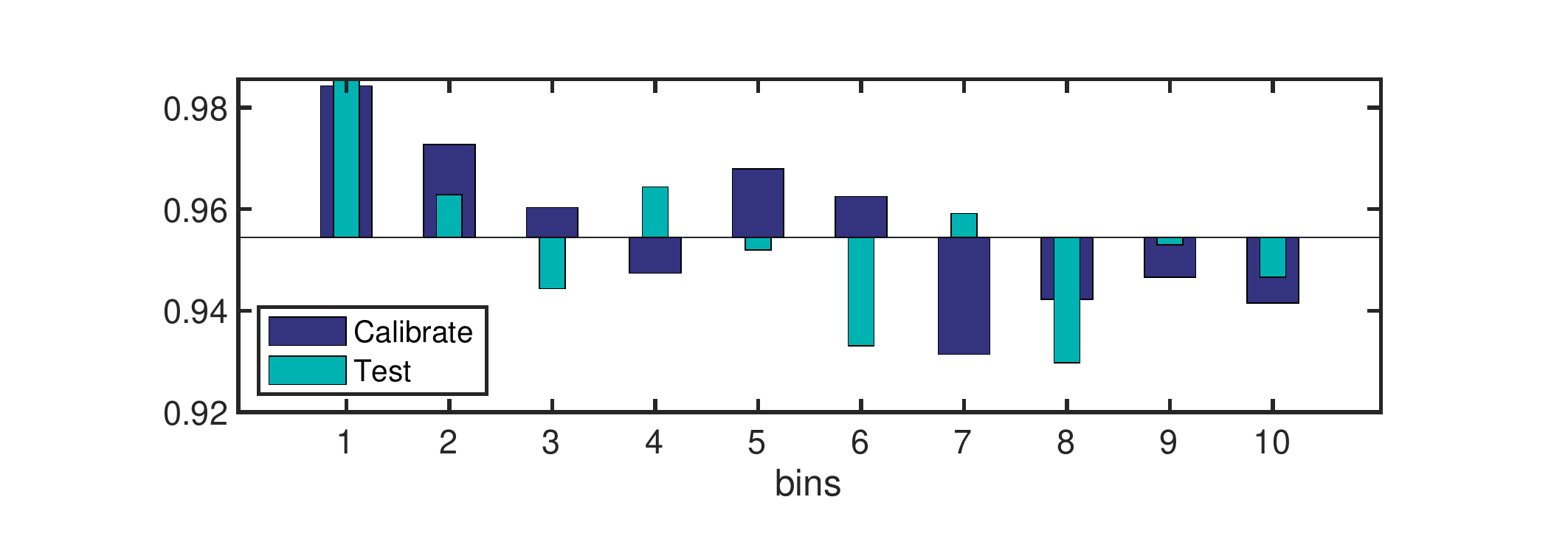}\includegraphics[width=.33\paperwidth]{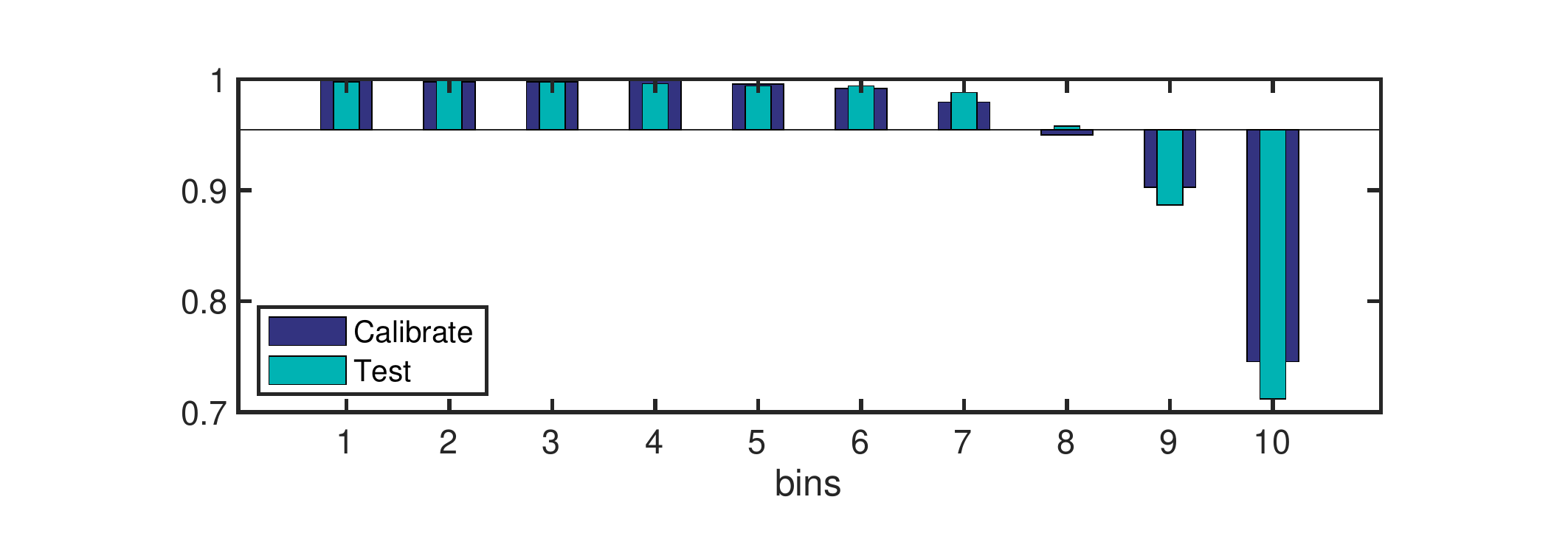}   }
\caption{Confidence bins on CIFAR-10:  PMax, Human classification,  Dropout .005, Dropout .02. The first dropout is not effective: the bins are small and poorly calibrated.} 
\label{fig33}
\end{center}
\vskip -0.2in
\end{figure*}

The scores are presented in Table~\ref{tab:CIFAR10} and visualized in \autoref{fig33}.
 We compared results using 10 bins and 100 bins.  These results confirm the notion that our models are calibrated, up to error coming from bin sampling, which is small.   The calibration error is small for all models, with ECE of $0.0096 \pm  0.0029$ for 10 bins, and ECE at most $.020 \pm .002$ for 100 bins.  
 
 The scores were consistent going from 10 to 100 bins. The EOR values increase going from 10 to 100 bins, which is expected, since we are resolving more fine grained probabilities.  For example, the Vanilla method went from 5.5 to 26 on EOR.   Dropout .02, which has the highest EOR value, was not resolved at 10 bins: the relative deviation in the EOR was order one. However, at 100 bins, the score stabilized to the high value of 32 with standard deviation of 1.5.

Across both 10 and 100 bins, the Brier score ranks Vanilla best, the NLL score ranks (Vanilla) and Dropout .2 equally.   According to EOR, the Dropout .2 is best, but the variance is large for 10 bins.  At 100 bins, the variance of EOR is small.   The difference in the scores is explained by looking at the bin probabilities. For Pmax the first five bins have average value of 0.9886, with corresponding odds ratio of 87.  For Dropout, the first five bins have an average value: 0.9983, with corresponding odds ratio of 592. It is the higher probabilities in the Dropout bins that contribute to the higher EOR score.  Thus, in the regime of very accurate models, with large bins, we should use EOR to make fine-grained distinctions between calibrated predictive uncertainty measures.

 \begin{table*}
  \caption{Scoring on CIFAR-10 using 10 bins (and 100 bins)}
   \label{tab:CIFAR10}
\begin{center}
\begin{small}
\begin{sc}
    \begin{tabular}{ l c c c}
        \toprule
      			& Brier  		& NLL & EOR   \\ \hline
    Vanilla    	& \textbf{0.0343}   & \textbf{0.13} & 5.51 $\pm .8$  \\ 
    Human  		& 0.0394 		& 0.16   &  3.02 $\pm .4$  \\
    Dropout .005 & 0.0436 		&  0.18 & 1.42 $\pm .01$   \\
    Dropout .02  &   0.0378 	&  \textbf{0.13}  &   $\mathbf{98.7 \pm 66}$  \\
    \hline
       Vanilla  (100 bins)  	& \textbf{0.0343}   & \textbf{0.14} & ${26 \pm 3}$  \\ 
    Human (100 bins) 		& 0.0406 		& 0.16   &  4 $\pm 1.4$  \\
    Dropout .005 (100 bins)& 0.0445 		&  0.20 &  $7\pm 1$   \\
    Dropout .02  (100 bins)&   0.0372 	&  \textbf{0.14}  &   $\mathbf{32 \pm 1.5}$  \\

     \end{tabular}
 \end{sc}
\end{small}
\end{center}
\end{table*}

\subsection{Top-1 and Top-5 uncertainty on ImageNet}
\begin{figure*}
\vskip 0.2in
\begin{center}
  \centering
   {\includegraphics[width=.31\paperwidth]{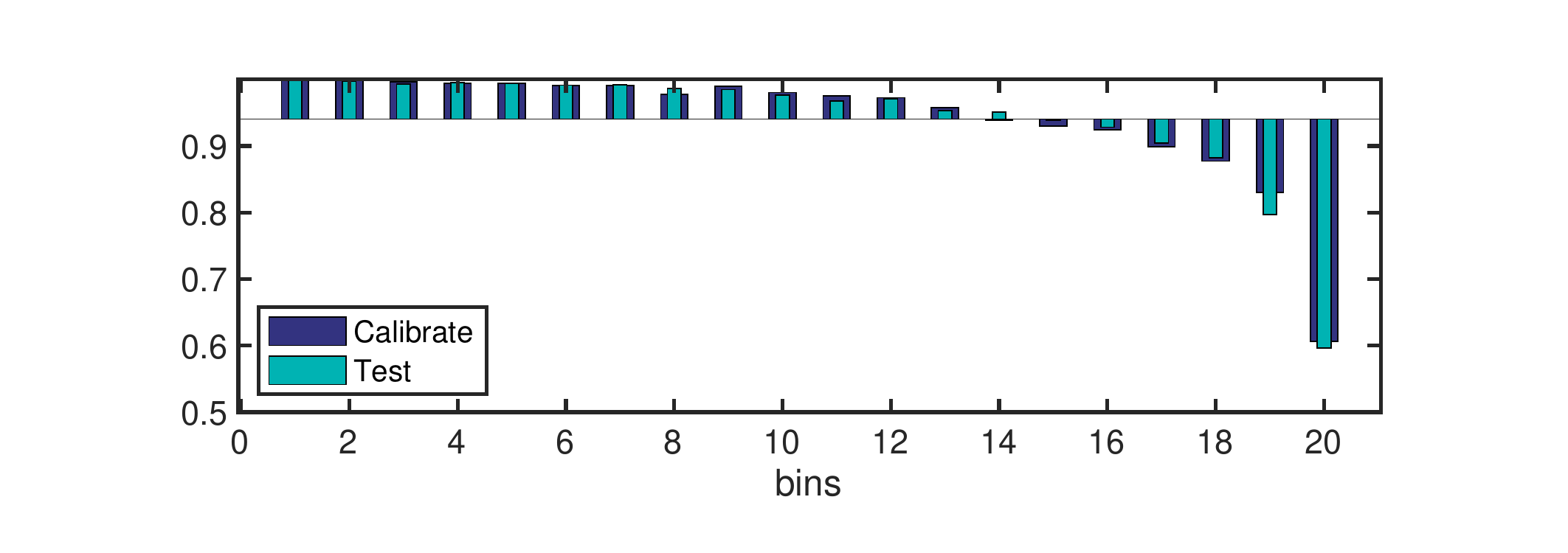}\includegraphics[width=.31\paperwidth]{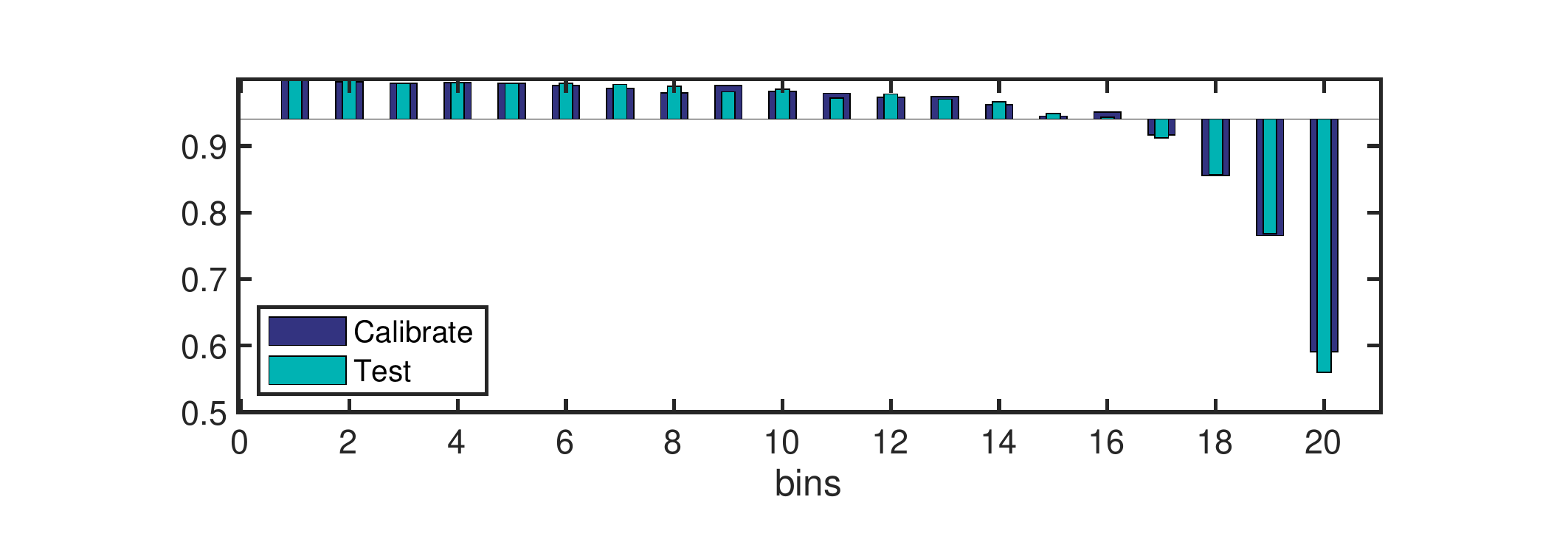}}
  { \includegraphics[width=.31\paperwidth]{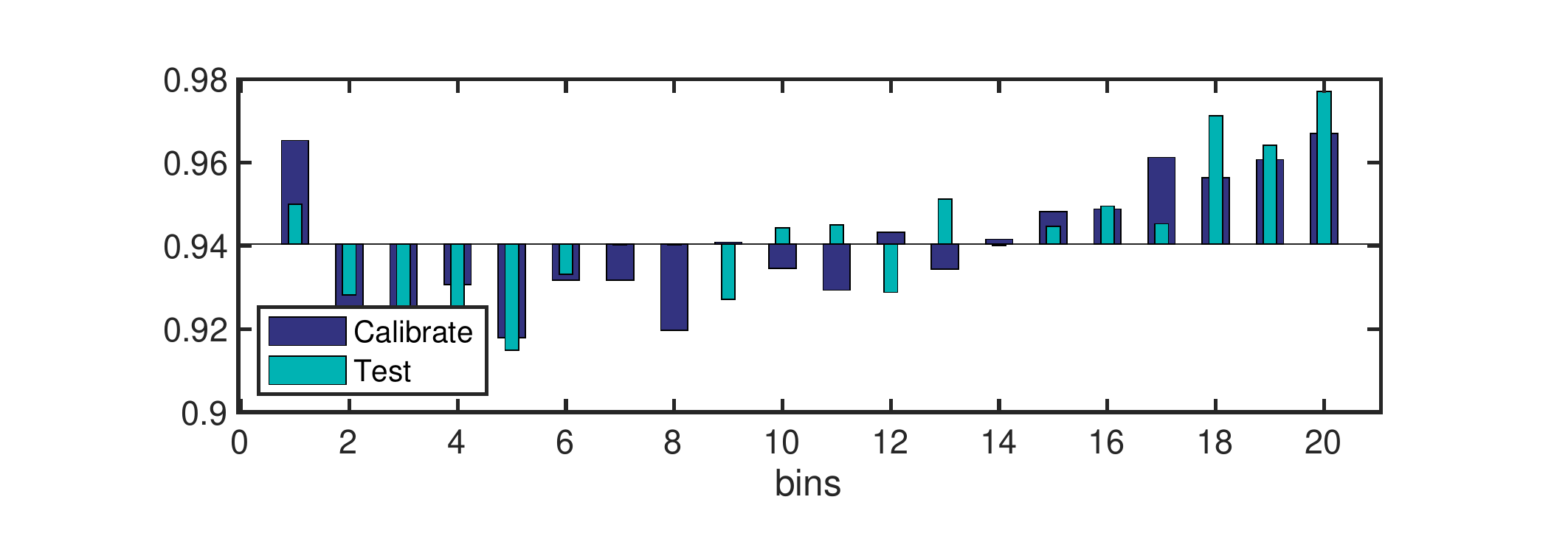}\includegraphics[width=.31\paperwidth]{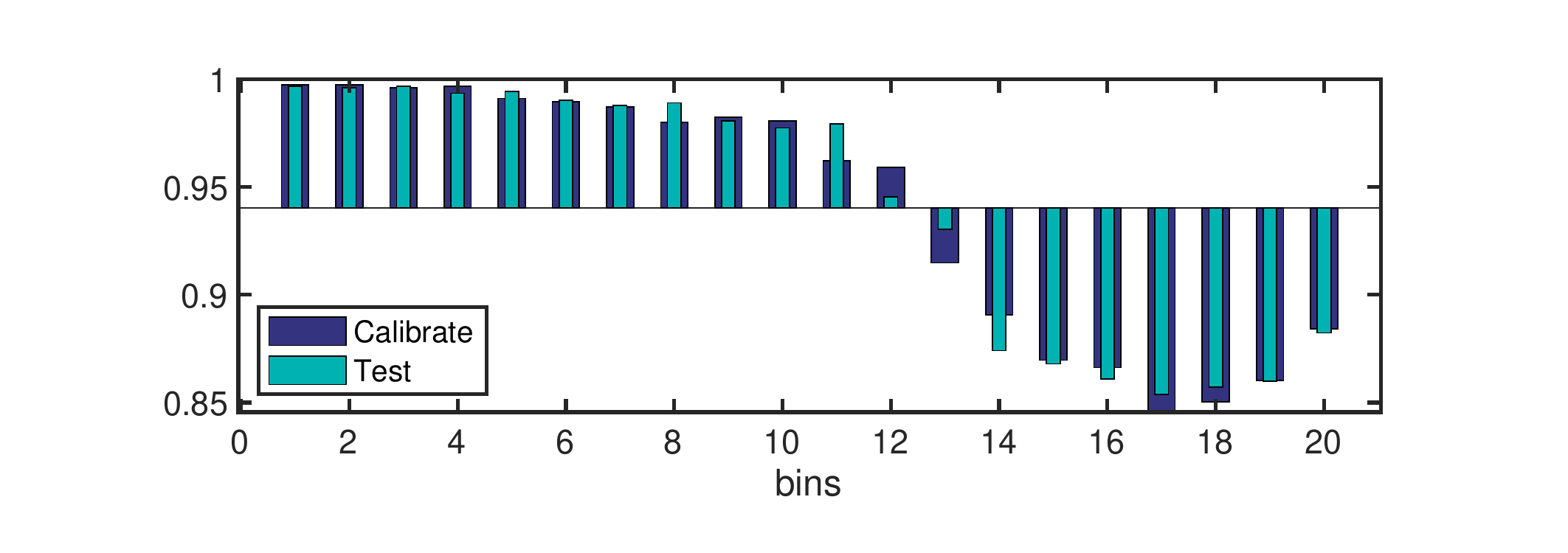}   }
     {\includegraphics[width=.31\paperwidth]{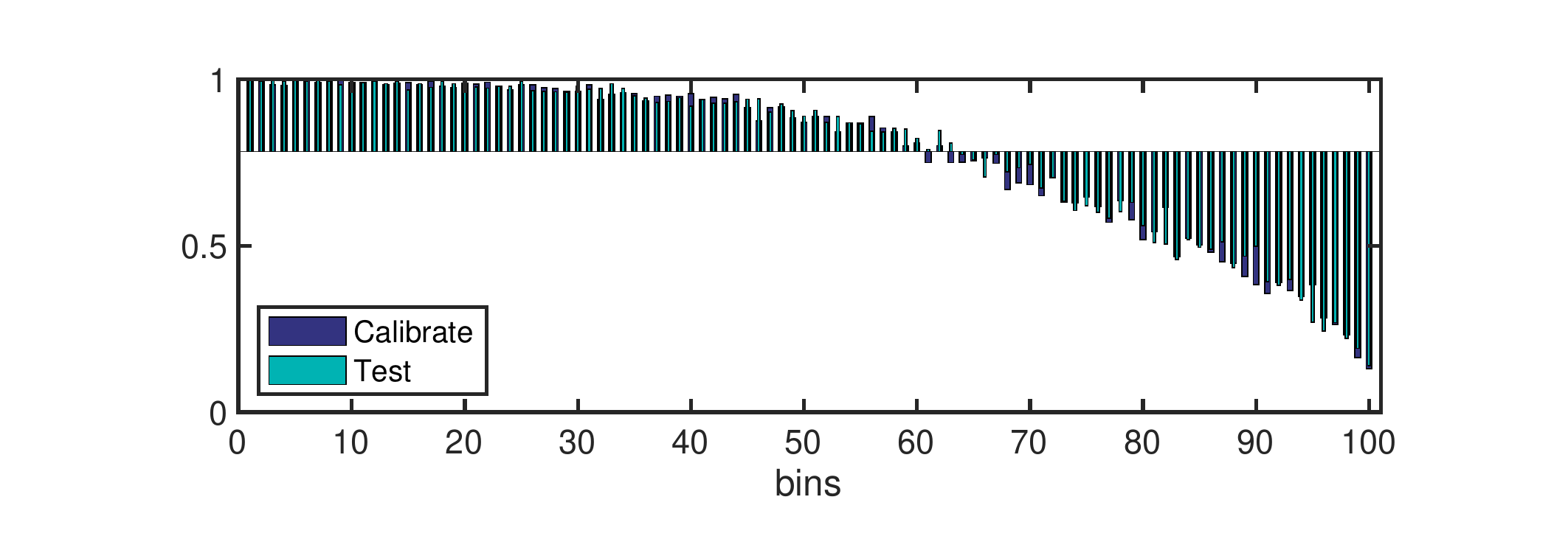}\includegraphics[width=.31\paperwidth]{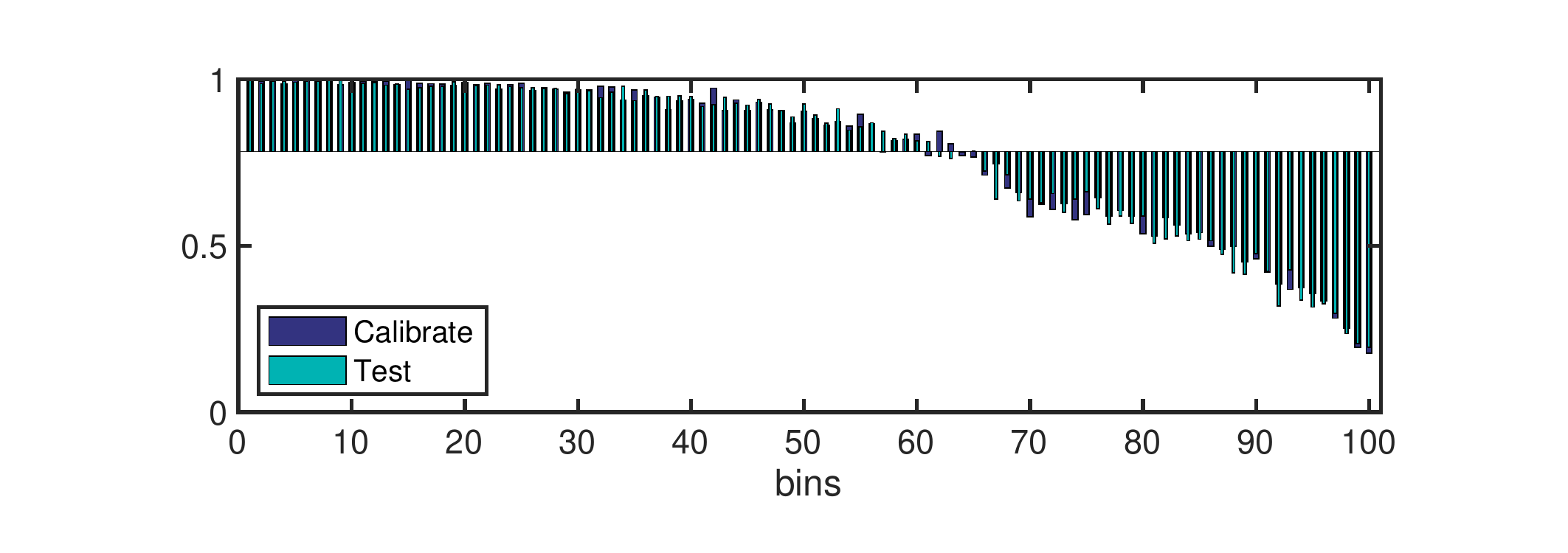}}
  { \includegraphics[width=.31\paperwidth]{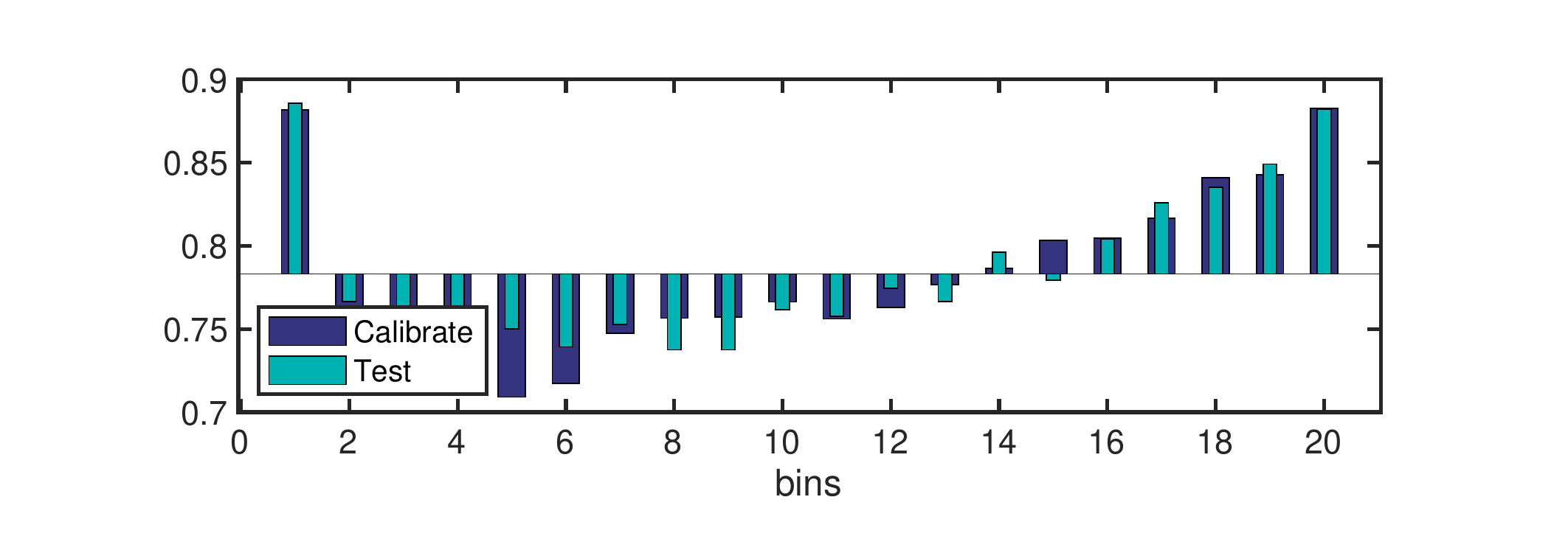}\includegraphics[width=.31\paperwidth]{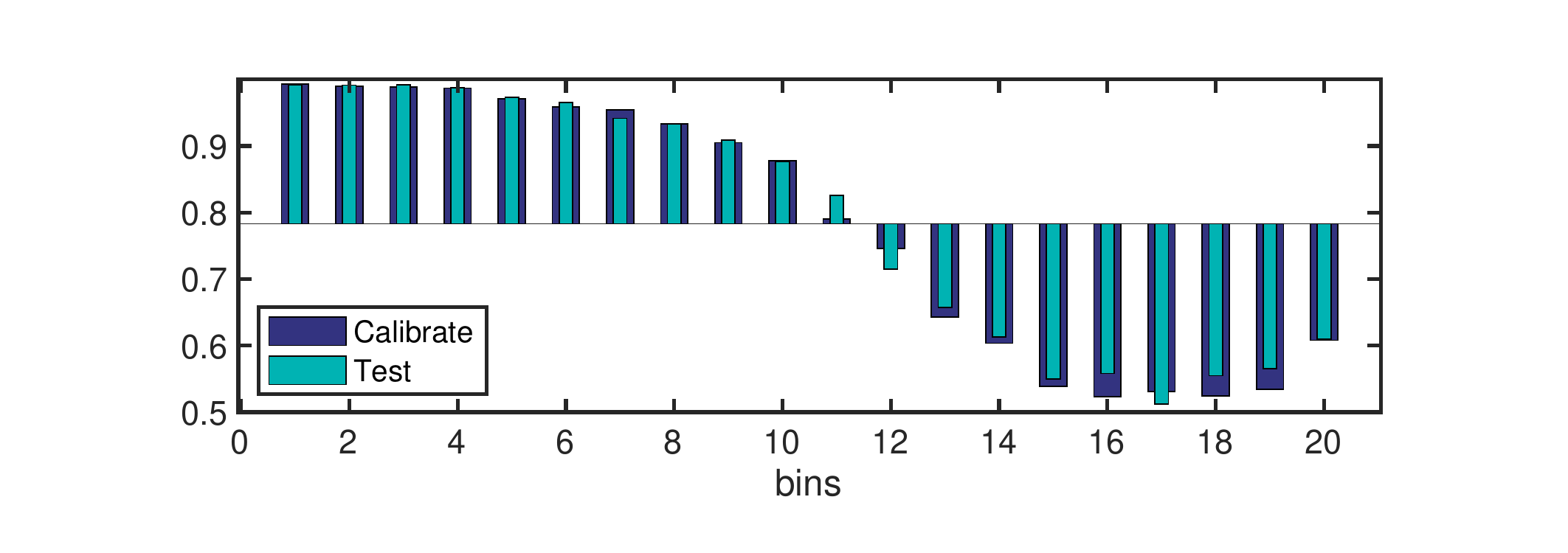}   }  
\caption{ Top-5 (first four) and Top-1 (next four) Confidence bins on ImageNet. In order from top left to bottom right: Vanilla, Entropy, Dropout .02, Dropout .005.  Note the shape of the bins is similar, but the width depends on the accuracy. } 
\label{figIMTOP5}
\end{center}
\vskip -0.2in
\end{figure*}

In this section, we score the Top-5 and Top-1 accuracy on ImageNet, comparing results with 20 and 100 bins.  See Tables~\ref{tab:Image22} and~\ref{tab:I1} and \autoref{figIMTOP5}.

Calibration:  the ECE was $.01 \pm .002$ ($.02 \pm .002$) for three of the measures, and $0.012 \pm .002$ ($0.03 \pm .002$) for the less effective dropout, using 20 and 100 bins, respectively.
The Brier and ECE scores were consistent across both 20 and 100 bins, which suggests that the increase in resolution was balanced by the decreased calibration.   The EOR increased when the number of bins was increased, as expected, since it is a measure of spread that does not take into account calibration.

For Top-1 accuracy, Vanilla was the best method, according to Brier, NLL, and EOR.
Dropout .005 was the lease effective measure, as illustrated by the EOR, which is close to the minimum of 1. For Top-5 accuracy, Entropy comes out at top ranked by both Brier and NLL.  However, the EOR spread is better for Vanilla with 10 bins, and nearly equal for Vanilla and Entropy with 100 bins. 
It is reasonable the Vanilla worked best for Top-1, since it corresponds to PMax, likewise, entropy also takes into account the lower ranked probabilities, so it seems reasonable that it works better for Top-5.  

In terms of scoring methods, NLL had a wider relative range of values, compared to the Brier score, so it made larger distinction between methods, but the ranking of the methods is the same.  The EOR values can have large variance with small bins, however the wide range of scores emphasizes distinctions between the models.

 \begin{table*}
\caption{Scoring Top-1 on ImageNet using 20 bins (100 bins)}
   \label{tab:I1}
\begin{center}
\begin{small}
\begin{sc}
    \begin{tabular}{ l c c c}
        \toprule
        & Brier  & NLL & EOR   \\ \hline
    Vanilla    &   \textbf{0.115}           & \textbf{0.358}  &  $\mathbf{13 \pm 2}$     \\ 
    Entropy &  0.117 	&   0.365   &  {12} $\pm 1$  \\
    Dropout .005  &    0.168 &  0.518  &  1.3 $\pm .02$  \\
    Dropout .02     &     0.136  &  0.408   &      8.9 $\pm 0.8$   \\  
   \hline 
      Vanilla (100 bins) &  \textbf{0.116}    &   \textbf{0.361}  &    $\mathbf{21\pm 2}$     \\
        Entropy (100 bins)		 & {0.118}    & {0.368}   & $ {   20 \pm 3}$   \\
           Dropout .005 (100 bins) &    0.168 &  0.516  &  1.5 $\pm .02$  \\ 
         Dropout .02   (100 bins)  &    {0.136}  &  0.411   &      ${   14 \pm 4}$        
     \end{tabular}
 \end{sc}
\end{small}
\end{center}
\end{table*}

\begin{table*}
  \caption{Scoring Top-5 on ImageNet using 20 bins (100 bins)}
   \label{tab:Image22}
\begin{center}
\begin{small}
\begin{sc}
    \begin{tabular}{ l c c c}
        \toprule
        & Brier  & NLL & EOR   \\ \hline
    Vanilla    &  0.048         & 0.175  &  $\mathbf{ 20 \pm  20}$    \\ 
    Entropy & \textbf{0.046}  	& \textbf{0.166}   &  $9 \pm  1$  \\
    Dropout .005  &  0.056  &  0.223 &  $1.3 \pm .02$   \\
    Dropout .02     &    0.053  &  0.194  &    $7 \pm 1$   \\
   \hline 
      Vanilla (100 bins) &  0.047     &   0.176    &    $\mathbf{ 35 \pm 5}$    \\
        Entropy (100 bins) & \textbf{0.046}   & \textbf{ 0.168 }   &   $\mathbf{  32 \pm 5}$   \\
            Dropout .005  &  0.056  &  0.227 &  $2 \pm 1$   \\
           Dropout .02 (100 bins)    &    0.052  &  0.195  &    $27 \pm 6$   
     \end{tabular}
 \end{sc}
\end{small}
\end{center}

\end{table*}

%

\begin{ack}
This material is based upon work supported by the Air Force Office of Scientific Research under award number FA9550-18-1-0167 (A.O.) and by a grant from the Innovative Ideas Program of the Healthy Brains and Healthy Lives initiative (HBHL), McGill University (C.F. and A.O).

\end{ack}

\medskip

\small

\bibliographystyle{alpha}
\bibliography{Confident}

\newpage
\section{Supplementary Material}

\subsection{Comparison of the expected odds ratio and Brier score}
We compared the expected odds ratio (EOR) and Brier score for various uncertainty methods on benchmark image classification datasets.  Since our methods are well-calibrated, we can see if the indication of resolution coming from the EOR gives more information. 

In Table~\ref{tab:brier} and Table~\ref{tab:BR} we present the Brier scores (smaller is better) and EOR values (larger is better) for various uncertainty methods on CIFAR-10, CIFAR-100, and ImageNet-1K. For CIFAR-10 we used the probability of the correct label; for CIFAR-100 and ImageNet-1K we used the probability that the correct label is in the
  $\mathrm{Top 5}$.   

On ImageNet, the model based uncertainty metrics perform better than dropout variance and ensemble variance, which are more expensive to compute, with regards to both Brier score and EOR. On the other hand, on CIFAR-10, Brier score prefers model entropy and Pmax, whereas the EOR prefers ensemble variance and the best choice of dropout. Finally, on CIFAR-100, the rankings are different again : ensemble variance has the best Brier score, and comes in second for EOR.  Entropy, Pmax and the $-\log(\sum p_{1:5})$  are exactly equal for the Brier score, but differently ranked by EOR, with the third one being best for EOR.

\begin{table*}
  \caption{Brier score of various measures of uncertainty. Using 100 bins equal sized bins.}
   \label{tab:brier}
\begin{center}
\begin{small}
\begin{sc}
  \begin{tabular}{ l     c c c }    
    \toprule
      uncertainty measure &  CIFAR-10 & CIFAR-100 & ImageNet-1K \\
      \midrule
      Model Entropy                &  \textbf{0.033}  & 0.067  & 0.041 \\
      $-\log p_\textrm{max}$       &  \textbf{0.033}  & 0.067  & 0.042 \\
      $-\log \sum p_{1:5}$         &  -      & 0.067  & \textbf{0.040} \\
    Dropout ($p=0.002$) & 0.036  &0.074  & 0.047  \\
    Dropout ($p=0.01$)  & 0.04 & 0.075  &  0.048 \\
    Dropout ($p=0.05$)  & 0.043 &  0.076  & 0.049 \\
    Ensemble variance            & 0.040 &  \textbf{0.050} & 0.047  \\
    \midrule
    Loss & 0  & 0.029  & 0.019 \\
    \bottomrule
  \end{tabular}
  \label{tab:Brier}
 \end{sc}
\end{small}
\end{center}
\end{table*}

\begin{table}
  \caption{Expected odds ratio for various  uncertainty methods.  Using 100 bins equal sized bins.}  
  \label{tab:BR}
\begin{center}
\begin{small}
\begin{sc}
  \begin{tabular}{ l     c c c }    
    \toprule
      uncertainty method &  CIFAR-10 & CIFAR-100 & ImageNet \\
      \midrule
      Model Entropy                &  4.29  & 3.64 & 8.18 \\
      $-\log p_\textrm{max}$       &  4.22  & 3.77  & \textbf{8.87} \\
    $-\log \sum p_{1:5}$         & -      & \textbf{4.25}  & 8.45  \\
    Dropout ($p=.002$) & 10.39  & 3.11 & 6.84  \\
    Dropout ($p=.01$)  & 4.67 & 2.38  &  7.81 \\
    Dropout ($p=.05$)  & 1.69 &  1.35 & 1.60 \\
    Ensemble Variance           & \textbf{16.66} & 4.03 & 6.13 \\
    \bottomrule
  \end{tabular}
\end{sc}
\end{small}
\end{center}
\end{table}

\begin{figure*} 
 \includegraphics[width=0.95\textwidth]{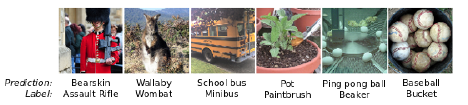}  
\caption{High uncertainty (low loss) images which are incorrectly classified.  These are mislabelled or ambiguous.}  
  \label{fig:mis-labeled}
\end{figure*}

\subsection{Finding mislabelled images}
We used Pmax to detect incorrectly labelled images, see~ \cite{reed2014training} and~\cite{pleiss2020identifying}.  We looked for the most confident images which were incorrectly classified, using the most confident bins.   It turned out that all of these were either incorrectly labelled, or were ambiguous images, see illustrations in Figure~\ref{fig:mis-labeled}.  For example, in the second image, the animal is a wallaby, not a wombat.  In the fourth image, a paintbrush is a kind of plant, but there is also a pot in the image.

\subsection{Metrics for discrimination: ROC and AUROC}\label{sec:AUROC}
The receiver operating characteristic curve (ROC)  a graphical plot that illustrates the diagnostic ability of a binary classifier system as its discrimination threshold is varied \cite{davis2006relationship}.
The ROC curve is created by plotting the true positive rate (TPR) against the false positive rate (FPR) at various threshold settings.   This plot is summarized in a single statistic, which is the area under the ROC curve (AUROC), which is in the interval $(0,1)$.  The AUROC was used as a score for uncertainty in~\cite{hendrycks_baseline_2017}

The AUROC score requires the uncertainty random variable to have ordered bins, and forces the decision threshold to combine the probabilities from each bin.  In example~\ref{ex:auroc1} we show that different orderings of the histogram lead to different AUROC values.
In example~\ref{ex:auroc2} we show that the AUROC can be the same, for two examples with very different expected odds ratios, in particular, the second histogram has a very high accuracy bin which is reflected in the EOR, but not as much in the AUROC.

\begin{example}[AUROC depends on ordering]\label{ex:auroc1}
Consider the coin toss example~\ref{ex:coin}.  In this case, with the increasing ordering of the probabilities 
the AUROC is .83.  On the other hand with the reverse ordering, the AUROC is .17, and it can take other values in between.
The expected odds ratio (EOR) score is 8, independent of the ordering. 
\end{example}

\begin{example}[same AUROC different BR]\label{ex:auroc2}
Consider two equally weighted histograms given by  $p = (.15, .4, .8)$, $q = (.4,  .5, .99)$. 
The averages are different: $\bar p = .45, \bar q = .63$.
The AUROC($p,w$) = .79, AUROC($q,w$) = .78.  EOR($p,w$) = 3.6, EOR($q,w$) = 21.
\end{example}

\end{document}